\documentclass[conference,a4paper]{IEEEtran}

\IEEEoverridecommandlockouts
\usepackage{cite}
\usepackage{amsmath,amssymb,amsfonts}
\usepackage{graphicx}
\usepackage{textcomp}
\usepackage{xcolor}
\usepackage{url}
\usepackage{algorithm}
\usepackage{booktabs}
\usepackage[noend]{algpseudocode}
\usepackage{longtable}
\usepackage{booktabs}
\usepackage{subcaption}
\usepackage{multirow}
\usepackage{diagbox}
\def\BibTeX{{\rm B\kern-.05em{\sc i\kern-.025em b}\kern-.08em
    T\kern-.1667em\lower.7ex\hbox{E}\kern-.125emX}}

\begin{document}

\title{A Conformal Predictive Measure for Assessing Catastrophic Forgetting}

\author{
    \IEEEauthorblockN{
        Ioannis Pitsiorlas\IEEEauthorrefmark{2}\textsuperscript{*}, Nour Jamoussi\IEEEauthorrefmark{2}\textsuperscript{*}, Marios Kountouris\IEEEauthorrefmark{2}\IEEEauthorrefmark{3}
    }
    \IEEEauthorblockA{\IEEEauthorrefmark{2}{Communication Systems Department, EURECOM, France}}
    \IEEEauthorblockA{\IEEEauthorrefmark{3}{Andalusian Institute of Data Science and Computational Intelligence (DaSCI)}\\
    \IEEEauthorrefmark{0}{Department of Computer Science and Artificial Intelligence, University of Granada, Spain}}
    \IEEEauthorblockA{\textsuperscript{*}\textit{Equal contribution}}
}

\maketitle

\begin{abstract}
This work introduces a novel methodology for assessing catastrophic forgetting (CF) in continual learning. We propose a new conformal prediction (CP)-based metric, termed the \textit{Conformal Prediction Confidence Factor (CPCF)}, to quantify and evaluate CF effectively. Our framework leverages adaptive CP to estimate forgetting by monitoring the model's confidence on previously learned tasks.
This approach provides a dynamic and practical solution for monitoring and measuring CF of previous tasks as new ones are introduced, offering greater suitability for real-world applications. Experimental results on four benchmark datasets demonstrate a strong correlation between CPCF and the accuracy of previous tasks, validating the reliability and interpretability of the proposed metric. Our results highlight the potential of CPCF as a robust and effective tool for assessing and understanding CF in dynamic learning environments.

\end{abstract}


\begin{figure*}[htb]
\centering
\includegraphics[scale=0.547]{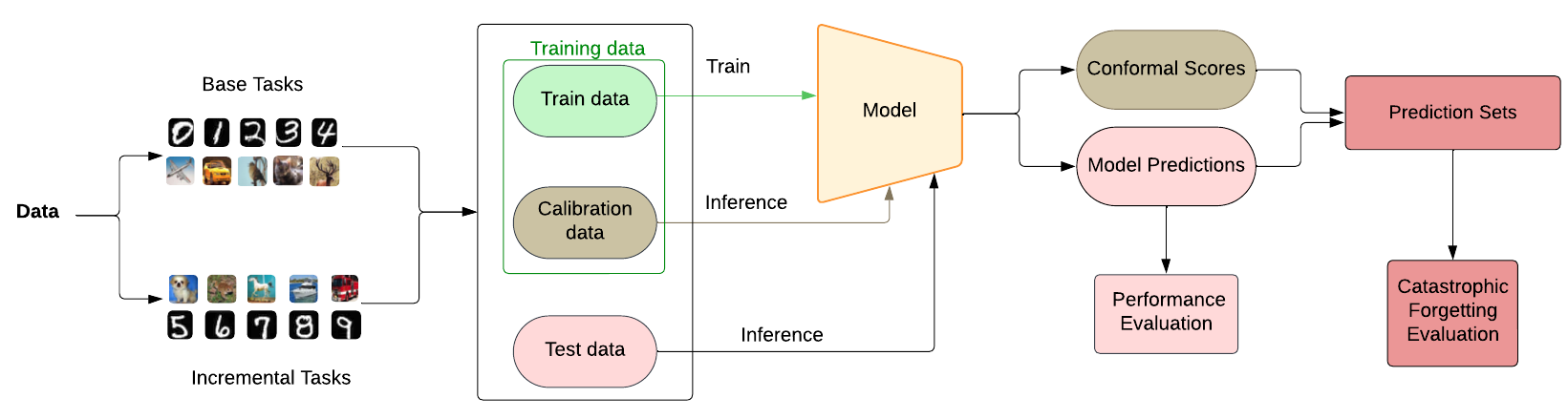}
\caption{Proposed framework overview for assessing catastrophic forgetting in continual learning via conformal prediction. 
}
\label{fig:frameowrk}
\end{figure*}

\section{Introduction}
Current machine learning (ML) models are predominantly designed for single-task applications, where they undergo iterative training over multiple epochs on a static dataset until convergence is reached \cite{serra2024leveragepredictiveuncertaintyestimates}. However, significant challenges arise when models are trained exclusively on fixed datasets, particularly in real-world scenarios where new tasks or classes may frequently emerge over time. For instance, the transition to 6G networks in wireless communications calls for adaptive resource allocation models that can accommodate new devices or services without requiring complete retraining. As user demands and network configurations change, traditional ML models often fail to generalize effectively, rendering full retraining costly and impractical in dynamic environments. A similar issue arises in healthcare, where ML models used for diagnostics are typically trained on known diseases but must also adapt to new conditions, such as emerging virus strains or rare diseases, without compromising accuracy on previously learned cases \cite{serra2024leveragepredictiveuncertaintyestimates}. These challenges highlight the need for ML models that can continuously learn and adapt to dynamic, real-world environments while preserving their performance on previously learned tasks.

To address this challenge, continual learning (CL), also known as lifelong or incremental learning, has gained considerable attention. CL aims to incrementally learn from a sequence of tasks or data points while preserving the performance of the model as if all data were available concurrently \cite{10444954}. A key requirement of a CL system is its ability to adapt to new tasks or shifts in data distribution without relying on access to previously encountered data \cite{nguyen2019understandingcatastrophicforgettingcontinual}.
This capability is particularly critical in scenarios where data privacy is a priority, such as under General Data Protection Regulation (GDPR) compliance, where organizations are obligated to restrict data usage and storage beyond a specified retention period. These restrictions significantly complicate the process of reusing historical datasets for retraining. Moreover, the high computational cost of retraining, especially for large-scale models, presents another major barrier when integrating new data \cite{shahid2025large}. By addressing both data privacy and computational efficiency, CL offers a promising solution for enabling models to evolve over time while remaining compliant with privacy regulations and minimizing resource demands.

Continual learning systems often suffer from catastrophic forgetting (CF), a phenomenon in which previously acquired knowledge is overwritten as new information is learned \cite{MCCLOSKEY1989109}. This typically occurs when a model adapts to a shift in the data distribution, resulting in a degradation in performance on earlier tasks or previously encountered distributions.

Distribution shifts, such as covariate, label, and concept shifts, refer to changes in the underlying data distribution encountered by a model during training or deployment, often leading to significant performance degradation. Covariate shift involves changes in the input distribution, while the relationship between the input and output remains stable \cite{8978471}. Label shift refers to changes in the distribution of output labels while the input distribution stays the same \cite{lipton2018detectingcorrectinglabelshift}. Concept shift refers to changes in the input-output relationship itself, which often arises when new tasks are introduced \cite{vovk2020testingconceptshiftonline}.

Over the years, numerous methods have been proposed in CL to address the challenge of CF. These approaches generally include regularization-based approaches, such as Elastic Weight Consolidation (EWC) \cite{Kirkpatrick_2017}, Online Structured Laplace Approximations \cite{ritter2018onlinestructuredlaplaceapproximations}, replay-based approaches, such as Reservoir Sampling \cite{reservoir}, and optimization-based strategies, such as Gradient Episodic Memory, among others \cite{wang2024comprehensivesurveycontinuallearning}. 
A significant limitation of current CL methods is their tendency to overlook predictive uncertainty, often leading to overconfident yet unreliable decisions \cite{einbinder2022traininguncertaintyawareclassifiersconformalized}. This limitation becomes particularly pronounced and problematic in real-world applications, where models encounter dynamically evolving tasks and data distributions. In high-stakes domains such as healthcare (e.g., risk of misdiagnosis), autonomous systems, or intrusion detection, overconfidence in predictions can result in critical errors, compromised safety, and severe real-world consequences \cite{ 10778374, 10641865}.

Quantifying predictive confidence is essential not only to improve decision-making, but also to effectively assess and mitigate CF. Traditional methods, such as softmax-based distillation, often produce overconfident predictions and fail to capture the underlying uncertainty in model parameters or data distributions \cite{kurmi2021forgetattenduncertaintymitigating}. In contrast, uncertainty-aware approaches, including those leveraging Bayesian techniques or ensemble methods, have shown promising potential. However, these methods are typically computationally intensive and difficult to scale in CL settings, where efficiency and adaptability are critical requirements \cite{kurmi2021forgetattenduncertaintymitigating}.

Conformal prediction (CP) has been successfully applied in various domains, including medical diagnostics, anomaly detection, and uncertainty quantification, to provide calibrated confidence measures for model predictions \cite{angelopoulos_intro}. As a model-agnostic and principled framework, CP offers an effective way to quantify uncertainty by providing calibrated confidence levels that remain reliable and can adapt to dynamic learning environments.

In this work, we propose a conformal predictive measure for assessing CF, termed the Conformal Prediction Confidence Factor (CPCF). By integrating CP into the CL pipeline, our approach enables an online evaluation framework for CF, enhancing the evaluation process to better reflect the challenges of dynamic real-world environments. Our empirical results demonstrate a strong correlation between the proposed CPCF and model accuracy, in two different learning settings and four benchmark datasets, underscoring its effectiveness in detecting CF.

\section{Proposed Methodology}
In this section, we introduce \textit{Adaptive CP} \cite{angelopoulos_intro} for classification tasks and propose a new metric, the CPCF, to assess CF in CL settings. An overview of the proposed framework is shown in Figure \ref{fig:frameowrk}.

\subsection{Data Splitting}
To implement CP, the training dataset is split into two subsets according to a predefined calibration ratio.

\begin{itemize} 
\item \textbf{Effective Training Data:} The portion of the dataset that is used to train the model. 
\item \textbf{Calibration Data:} The remaining portion of the dataset, specified by the calibration ratio, is reserved for computing conformal scores, which are essential for forming prediction sets during inference. 
\end{itemize}

For example, if the calibration ratio is set to $0.1$, then 90\% of the data is allocated for training while the remaining 10\% is used for calibration. Clearly, the calibration ratio influences the interplay between training and calibration, balancing predictive performance against the reliability of uncertainty quantification. Larger calibration ratios may enhance the reliability of conformal scores, but this comes at the expense of reducing the amount of data available for training. Conversely, smaller calibration ratios favor training by allocating more data, but may compromise the quality of the calibration, and consequently, the effectiveness of the prediction sets.

\subsection{Calibration Phase}
Let $\mathcal{X}_c = \{x_{c,i}\}_{i=1}^n$ denote the calibration set. The calibration phase is conducted to compute conformal scores and establish the adjusted quantile threshold $q_\alpha$. This process involves the following steps:
\begin{enumerate}
    \item For each calibration point $x_{c,i} \in \mathcal{X}_c$:
    \begin{itemize}
        \item Compute the model's output logits and apply the softmax function to obtain class probabilities.
        \item Rank the classes in descending order based on their softmax scores.
        \item Identify the rank of the true label $y_{c,i}$ within the sorted class probabilities.
        \item Compute the conformal score $E_i$, which represents the minimum cumulative probability mass required to include the true label:
        \begin{equation}
            E_i = \sum_{k=1}^{\text{rank}(y_{c,i})} \hat{\pi}(x_i)_{(y_k)}
        \end{equation}
        where $\hat{\pi}(x_i)_{(y_k)}$ denotes the softmax score of the $k$-th ranked in descending order class.
    \end{itemize}
    \item Determine the adjusted quantile threshold $q_\alpha$ using the set of computed conformal scores and a chosen significance level $\alpha$:
    \begin{equation}
        q_\alpha = \text{Quantile}(\{E_i\}_{i=1}^n, \frac{\lceil (n + 1)(1 - \alpha) \rceil}{n})
    \end{equation}
    where $n$ is the size of the calibration set.
\end{enumerate}

\subsection{Prediction Phase} 
Given a test point $x_t$, the prediction phase uses the previously computed adjusted quantile threshold $q_\alpha$ to form the CP set corresponding to the point $x_t$. The steps involved are as follows:
\begin{enumerate}
    \item Compute the model's output logits for $x_t$ and apply the softmax function to obtain class probabilities.
    \item Rank the classes in descending order based on their softmax scores and compute the cumulative sum of the probabilities.
   \item Construct the conformal prediction set $\mathcal{C}(x_t)$, which includes the most likely classes, i.e., the classes whose cumulative probability mass is bigger than or equal to the threshold $q_\alpha$:
    \begin{equation}
        \mathcal{C}(x_t) = \{y_k \mid k \in \{1, \dots, K\} , \sum_{k=1}^{K} \hat{\pi}(x_t)_{(y_k)} \geq q_\alpha\},
    \end{equation}
    where $\hat{\pi}(x_t)_{(y_k)}$ represents the softmax score of the $k$-th class ranked in descending order, and $q_\alpha$ is the quantile threshold.
\end{enumerate}

This process is repeated for each test sample, resulting in a CP set for every instance. The length of these prediction sets serves as a calibrated measure of confidence in the predictions of the model, allowing a comprehensive evaluation of its predictive performance and uncertainty.

\subsection{Proposed Metric: CPCF}
The key contribution of this work is the introduction of the CPCF as a novel metric to evaluate CF in CL settings using CP sets. 

The central idea is to leverage CP sets, which quantify the model's uncertainty, to evaluate how well the model retains knowledge from previously learned tasks after training on a new task. By examining the average length of prediction sets corresponding to previously learned tasks, the CPCF effectively captures the model's confidence in its prior knowledge. A shorter average prediction set indicates higher confidence and stronger retention, while longer sets suggest increased uncertainty and potential forgetting. 

Specifically, to compute the CPCF after training the model on task $j$, we follow the steps outlined in Algorithm \ref{alg:cpcf}. 

After training the model on the new task $j$, we use the calibration sets from all previous tasks, i.e., up to task $j-1$ to compute the corresponding conformal scores $\{E_i\}_{i=1}^{l_j}$, where $l_j=\sum_{k=1}^{j-1} n_k$ and $n_k$ denote the size of the calibration set for task $k$. Then, we determine the corresponding quantile threshold $q_\alpha^j$ to construct the prediction set $\mathcal{C}(x_t)$ for each test sample from the test sets of the previous tasks, i.e., up to task $j-1$. 
The length of each resulting prediction set is computed, and the CPCF is obtained as the average prediction set length across all test samples up to the previous task. This process quantifies the model's confidence in its prior knowledge and serves as a quantitative measure of CF of previous tasks after training the model on a new task $j$. 

The CPCF metric provides a quantifiable approach to assess CF by observing changes in the lengths of CP sets. Shorter prediction set lengths indicate higher confidence and better retention of prior knowledge, whereas longer lengths suggest increased uncertainty, potentially resulting from forgetting previously learned tasks.

\begin{algorithm}[t]
\caption{Computing CPCF after Training on a new Task}
\label{alg:cpcf}
\begin{algorithmic}[1]
\State \textbf{Input:} Trained model up to task $j-1$, training data for task $j$, calibration set for tasks $1$ to $j-1$, i.e., $\mathcal{X}_c^{1:j-1}$, and test data for tasks $1$ to $j-1$, i.e., $\mathcal{X}_t^{1:j-1}$.
\State \textbf{Output:} $\text{CPCF}_{j}$ assessing the CF of previous tasks, i.e., tasks $1,2, \dots, j-1$.

\State \textbf{Step 1: Train the Model on Task $j$}

\State \textbf{Step 2: Compute Conformal Scores of Previous Tasks Calibration Samples}
\For{each calibration sample $x_{c,i} \in \mathcal{X}_c^{1:j-1}$}
    \State Compute the conformal score $E_i$.
\EndFor

\State \textbf{Step 3: Compute the Quantile Threshold $q_\alpha^j$}
\State Determine the quantile threshold $q_\alpha^j$:
\[
q_\alpha^j = \text{Quantile}(\{E_i\}_{i=1}^{l_j}, \frac{\lceil (l_j + 1)(1 - \alpha) \rceil}{l_j}),
\]
where $l_j = |\mathcal{X}_c^{1:j-1}|$
\State \textbf{Step 4: Form Prediction Sets of Previous Tasks Test Samples}
\For{each test sample $x_t \in \mathcal{X}_t^{1:j-1}$}
    \State Form the prediction set $\mathcal{C}(x_t)$
\EndFor

\State \textbf{Step 5: Compute CPCF assessing CF of Tasks $1, \dots, j-1$}
\[
\text{CPCF}_{j} = \frac{1}{|\mathcal{X}_t^{1:j-1}|} \sum_{x_t \in \mathcal{X}_t^{1:j-1}} |\mathcal{C}(x_t)|,
\]
\State \textbf{Return:} $\text{CPCF}_{j}$.
\end{algorithmic}
\end{algorithm}

\section{Experimental Results}
\subsection{Datasets Used}
We evaluate our approach on four benchmark datasets: MNIST, CIFAR-10, KMNIST, and FashionMNIST. These datasets encompass a diverse set of visual classification tasks, ranging from simple digit recognition (MNIST, KMNIST) to more complex object recognition (CIFAR-10, FashionMNIST), and include both grayscale and color images.

\subsection{Model Architecture}
We employ a three-layer Multi-Layer Perceptron (MLP) for all experiments. It processes flattened image inputs, with a dimension of $784$ for MNIST, KMNIST, and FashionMNIST, and $1024$ for CIFAR-10.
The network consists of two hidden layers with $256$ and $128$ ReLU-activated neurons, respectively. These are followed by a 10-neuron output layer (one neuron per class), with a softmax activation function applied to generate class probabilities.

\subsection{Training Curriculum}
During training, the model learns a mapping $f_\theta: \mathcal{X} \to \mathcal
{Y}$, where $\theta$ denotes the model parameters. Taking into account the CL setting based on incremental learning \cite{measuring_Kemker}, the training process occurs in two main phases.
\begin{enumerate}
    \item \textbf{Base Training:} The model is initially trained for 8 epochs in the base task, which involves learning to classify the first set of classes $\{0, 1, 2, 3, 4\}$. This phase establishes a foundational representation of the data.
    \item \textbf{Incremental Training:} The model is subsequently updated incrementally to learn new tasks, with $3$ epochs allocated to learning each of the remaining $5$ classes. During this phase, the model aims to incorporate new knowledge while retaining previously learned information, thereby mitigating CF.
\end{enumerate}

The extended base training period, relative to the shorter incremental task training phases, enables the model to build a robust foundational representation for the initial set of tasks. In contrast, the reduced number of epochs during incremental learning highlights the challenges of CL, where the model must efficiently adapt to new tasks while simultaneously preserving the knowledge of previously learned ones.

\begin{figure}[H]
    \centering
    \includegraphics[width=0.78\linewidth]
    {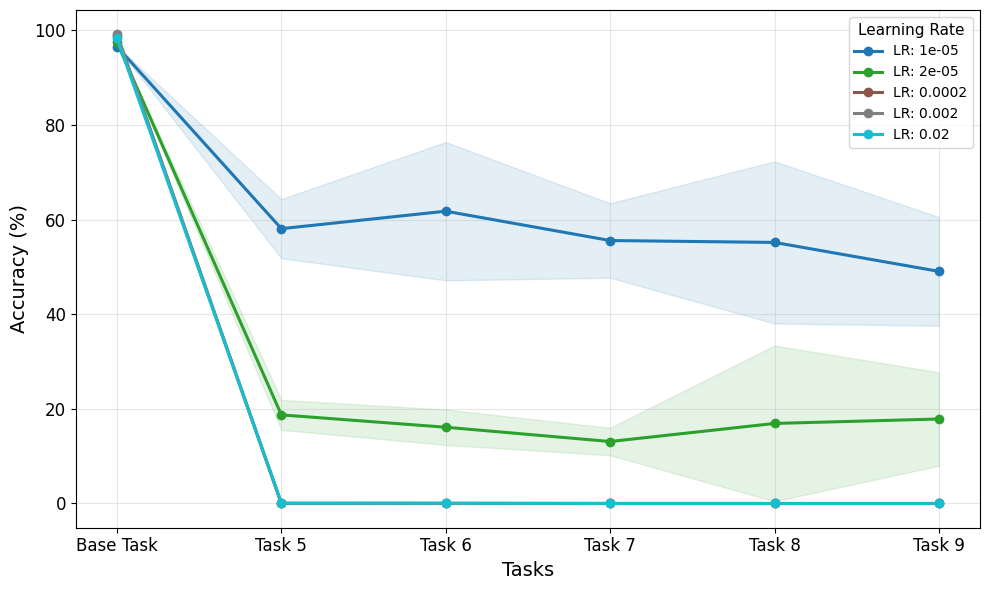}
    \caption{Impact of learning rate on CF in MNIST. The y-axis represents the accuracy on previously learned tasks.
}
    \label{fig:lr}
\end{figure}

The learning rate was set at $2 \times 10^{-5}$, a conservative choice aimed at balancing rapid adaptation with long-term retention. This training setup aligns with the principle \emph{“learn fast, forget fast; learn slow, forget slow”}, as illustrated in Figure~\ref{fig:lr}, where a slower learning rate helps mitigate CF. By adopting a gradual learning pace, the model retains prior knowledge more effectively while steadily integrating new information.

\subsection{Evaluation Metrics} 
To assess the model's ability to retain knowledge from previous training sessions, we adopt evaluation metrics inspired by \cite{measuring_Kemker}, which provide a robust baseline for evaluating CF.
We begin by defining the two key metrics used throughout this study.
\begin{itemize}
    
    \item \textbf{Accuracy on previously learned tasks} ($a_{\text{prev}}$): The average test accuracy over all tasks encountered prior to the current one. Specifically, $a_{\text{prev,}j}$ denotes the average test accuracy on tasks $1$ to $j{-}1$, measured after training on task $j$.

    \item \textbf{Accuracy on newly learned tasks} ($a_{\text{new}}$): The test accuracy on the most recently learned task $j$.
\end{itemize}
 
To show the effectiveness of the proposed CPCF in detecting and quantifying CF, we compare it with $a_{\text{prev}}$, and empirically show a strong correlation between CPCF and $a_{\text{prev}}$.

\subsection{Numerical Results}
By analyzing the relationship between CPCF and $a_\text{prev}$, we demonstrate the effectiveness of the metric in capturing the dynamics of forgetting previously learned knowledge. In addition, we investigate the robustness of CPCF to hyperparameter variations, showcasing its versatility across diverse learning scenarios. 
To further evaluate the reliability of CPCF in detecting CF, we compare results across two training paradigms: a standard MLP trained with cross-entropy loss, and the same MLP model trained with the EWC loss. 
By comparing CPCF's correlation with $a_{\text{prev}}$ in both settings, we assess whether CPCF remains consistent and sensitive in the presence of a forgetting mitigation strategy.

\begin{figure*}[htb]
\centering
\includegraphics[width=0.77\textwidth]{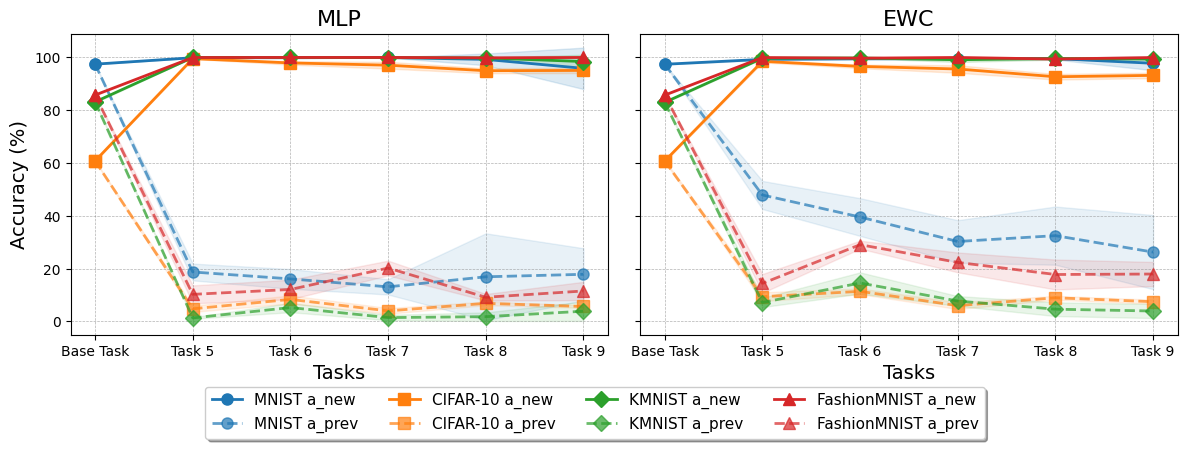}
\caption{Illustration of Catastrophic Forgetting: Comparison of accuracy on previously learned tasks ($a_{\text{prev}}$) and newly learned tasks ($a_{\text{new}}$) across incremental tasks.}
\label{fig:plot}
\end{figure*}

We evaluated two variants of the EWC to explore how different regularization strategies influence CF and model confidence.
The core idea behind EWC is to mitigate CF by selectively constraining the parameters that are most important to previously learned tasks. After training on a task \( A \), the model estimates the importance of each parameter by computing the diagonal of the Fisher Information Matrix \( F \), and stores the corresponding optimal parameters \( \theta^*_A \). This leads to the following regularized loss function when training on a subsequent task \( B \):
\begin{equation}
\mathcal{L}(\theta) = \mathcal{L}_B(\theta) + \frac{\lambda}{2} \sum_i F_i (\theta_i - \theta^*_{A,i})^2.
\end{equation}

This variant, which we refer to as \textit{single-penalization EWC}, applies a quadratic penalty that anchors the model parameters toward the solution found for task \( A \). The hyperparameter \( \lambda \) controls the strength of this regularization, balancing stability (retention of prior knowledge) and plasticity (adaptation to new tasks). 

When multiple tasks are introduced incrementally, the standard single-penalization EWC tends to primarily emphasize the most recently learned task. Thus, a \textit{multi-penalization EWC} variant that explicitly regularizes the model with respect to all previously learned tasks could be more effective for mitigating CF. This is achieved by maintaining a separate Fisher information matrix $F^{(j)}$ and the corresponding optimal parameters $\theta^*_j$ for each previous task $j$. The resulting loss function for training on task $t$ incorporates multiple regularization terms, one for each previous task:
\begin{equation}
\mathcal{L}(\theta) = \mathcal{L}_t(\theta) + \sum_{j=1}^{t-1} \frac{\lambda_j}{2} \sum_i F^{(j)}_i (\theta_i - \theta^*_{j,i})^2.
\end{equation}

Each term in the summation serves as an individual constraint aimed at preserving knowledge from a specific past task. This formulation offers stronger protection against forgetting by maintaining proximity to multiple past optima, though at the cost of increased memory requirements. To balance the influence of earlier tasks, we set $\lambda_j = \frac{\lambda}{2^{t-j-1}}$, assigning progressively greater weight to more recent tasks. This decay reflects the intuition that each task implicitly carries knowledge from earlier ones, thereby reducing the need to equally penalize distant past tasks.

To compare the two variants, we adopt the evaluation framework of \cite{measuring_Kemker}, which introduces the following three metrics: 
\begin{itemize}
    \item $\Omega_{\text{new}}$: the model’s average ability to learn and retain new tasks immediately after training,
    \item $\Omega_{\text{all}}$: the average accuracy across all learned tasks at each step, normalized by the base accuracy ($a_{\text{ideal}}$),
    \item $\Omega_{\text{base}}$: the retention of the base task after each new task is learned, normalized by $a_{\text{ideal}}$.
\end{itemize}

We also introduce \textbf{$\Omega_{\text{prev}}$}, which assesses the model’s ability to retain previously learned tasks after learning a new one. We compute it from the accuracy of previous tasks in each step, using $a_{\text{prev}}$. Formally, it is defined as:
\begin{equation}
\Omega_{\text{prev}} = \frac{1}{T - 1} \sum_{j=2}^{T} \frac{a_{\text{prev},j}}{a_{\text{ideal}}}.
\end{equation}

\begin{table}[h!]
\centering
\caption{Comparison of two EWC variants on CIFAR-10 using $\Omega$ metrics.}
\label{tab:ewc_comparison}
\begin{tabular}{lcccc}
\toprule
\textbf{Method} & $\Omega_{\text{base}}$ & $\Omega_{\text{new}}$ & $\Omega_{\text{all}}$ & $\Omega_{\text{prev}}$ \\
\midrule
Single-penalization EWC & 0.082 & 0.953 & 0.101 & 0.342 \\
Multi-penalization EWC & 0.082 & 0.953 & 0.101 & 0.342 \\
\bottomrule
\end{tabular}
\end{table}

As shown in Table~\ref{tab:ewc_comparison}, both EWC variants achieve identical scores on CIFAR-10 in all evaluation metrics: $\Omega_{\text{new}}$, $\Omega_{\text{all}}$, $\Omega_{\text{base}}$, and $\Omega_{\text{prev}}$. Although the single-penalization approach only penalizes deviations from the most recently learned task, it still indirectly preserves earlier knowledge, as each updated parameter configuration reflects the cumulative training history. Therefore, to reduce computational overhead and memory requirements, we adopted the single-penalization variant, referred to simply as EWC throughout the remainder of this study. 

Figure~\ref{fig:plot} highlights the phenomenon of CF, where the models studied maintain high accuracy in newly introduced tasks while exhibiting a sharp decline in accuracy performance in previously learned tasks. This drop in $a_{\text{prev}}$ reflects the model's difficulty in retaining knowledge from earlier tasks, ultimately leading to a failure to generalize across all learned tasks. The right panel shows that EWC partially alleviates this degradation by preserving higher $a_{\text{prev}}$ scores compared to standard MLP, particularly during the early stages of CL.

\begin{table*}[h!]
\centering
\begin{minipage}{0.48\textwidth}
\centering
\caption{Distance correlation between CPCF and $a_{\text{prev}}$ for fixed alpha ($\alpha=0.1$).}
\resizebox{\linewidth}{!}{%
\begin{tabular}{>{\centering\arraybackslash}p{1.2cm} >{\centering\arraybackslash}p{0.9cm} 
>{\centering\arraybackslash}p{1.3cm} >{\centering\arraybackslash}p{1.6cm} 
>{\centering\arraybackslash}p{1.8cm} >{\centering\arraybackslash}p{1.3cm}}
\toprule
\textbf{Calib.} & \textbf{Model} & \textbf{MNIST} & \textbf{CIFAR-10} & \textbf{FashionMNIST} & \textbf{KMNIST} \\ \midrule
0.05 & MLP & 0.5165 & 0.3009 & 0.2319 & 0.3214 \\
     & EWC & 0.7475 & 0.3848 & 0.2993 & 0.5970 \\ \midrule
0.10 & MLP & 0.5585 & 0.3211 & 0.2118 & 0.3121 \\
     & EWC & 0.6754 & 0.3883 & 0.3046 & 0.5996 \\ \midrule
0.15 & MLP & 0.5096 & 0.3363 & 0.1863 & 0.2709 \\
     & EWC & 0.7081 & 0.4414 & 0.3149 & 0.5693 \\ \midrule
0.20 & MLP & 0.4996 & 0.3319 & 0.1956 & 0.2655 \\
     & EWC & 0.7127 & 0.4834 & 0.2903 & 0.5298 \\ \bottomrule
\end{tabular}}
\label{tab:mlp_ewc_results}
\end{minipage}
\hfill
\begin{minipage}{0.48\textwidth}
\centering
\caption{Distance correlation between CPCF and $a_{\text{prev}}$ for fixed calibration ratio ($\text{calibration ratio} = 0.1)$.}
\resizebox{\linewidth}{!}{%
\begin{tabular}{>{\centering\arraybackslash}p{1.2cm} >{\centering\arraybackslash}p{0.9cm} 
>{\centering\arraybackslash}p{1.3cm} >{\centering\arraybackslash}p{1.6cm} 
>{\centering\arraybackslash}p{1.8cm} >{\centering\arraybackslash}p{1.3cm}}
\toprule
\textbf{Alpha} & \textbf{Model} & \textbf{MNIST} & \textbf{CIFAR-10} & \textbf{FashionMNIST} & \textbf{KMNIST} \\ \midrule
0.05 & MLP & 0.5347 & 0.2871 & 0.2611 & 0.2321 \\
     & EWC & 0.6250 & 0.2574 & 0.2740 & 0.5812 \\ \midrule
0.10 & MLP & 0.5585 & 0.3211 & 0.2118 & 0.3121 \\
     & EWC & 0.6754 & 0.3883 & 0.3046 & 0.5996 \\ \midrule
0.15 & MLP & 0.5732 & 0.3202 & 0.1834 & 0.3116 \\
     & EWC & 0.7480 & 0.4270 & 0.3073 & 0.6106 \\ \midrule
0.20 & MLP & 0.5919 & 0.3224 & 0.1777 & 0.3107 \\
     & EWC & 0.8012 & 0.4540 & 0.3157 & 0.6201 \\ \bottomrule
\end{tabular}}
\label{tab:distance_correlation_results}
\end{minipage}
\end{table*}

Tables~\ref{tab:mlp_ewc_results} and~\ref{tab:distance_correlation_results} reveal a strong distance correlation between CPCF and $a_{\text{prev}}$ in various calibration ratios and significance levels $\alpha$. Specifically, this correlation becomes consistently stronger when EWC is applied. This observation suggests that CPCF becomes more sensitive to forgetting in the presence of CF mitigation techniques, reinforcing its value as a reliable proxy for measuring forgetting in CL scenarios.

Table~\ref{tab:mlp_ewc_results} shows the distance correlation between CPCF and $a_{\text{prev}}$ in varying calibration ratios, with $\alpha$ fixed at $0.1$. Although some fluctuations are observed, there is no consistent trend suggesting that increasing the calibration ratio systematically improves or worsens the correlation. In some cases, a higher calibration ratio leads to marginal gains; in others, the correlation remains stable or slightly decreases. This behavior suggests that CPCF is relatively robust to the choice of calibration ratio, i.e., its alignment with forgetting, as measured by $a_{\text{prev}}$, does not critically depend on precise calibration tuning. In all settings, models trained with EWC consistently show stronger correlations than those trained with standard methods, further validating the sensitivity of CPCF under forgetting-aware regularization.

Table \ref{tab:distance_correlation_results} further explores the effect of varying the significance level $\alpha$. As expected, increasing $\alpha$ leads to larger prediction sets and stronger correlations with $a_{\text{prev}}$, as CPCF becomes more sensitive to subtle changes in model uncertainty. This behavior aligns with the theoretical role of $\alpha$ in CP: it directly determines the quantile threshold used to construct prediction sets, thus influencing their size and informativeness. A higher $\alpha$ corresponds to a lower threshold, resulting in larger prediction sets that are more likely to contain the true label but may also be less precise. These wider sets more effectively capture the model’s uncertainty, especially in regions of the input space where the model is less confident. Consequently, CPCF becomes increasingly sensitive to subtle degradations in model performance, particularly those introduced by forgetting previously learned tasks. This sensitivity explains the observed increase in correlation with $a_{\text{prev}}$ at higher $\alpha$ values.
In general, this behavior highlights an important strength of CPCF: its adaptability to different levels of uncertainty tolerance. By tuning $\alpha$, practitioners can control the granularity with which forgetting is detected, making CPCF a versatile tool for evaluating CL performance.

\section{Discussion and Concluding Remarks}
The proposed method presents a novel application of CP to quantify CF in CL, introducing CPCF as an effective and interpretable online metric. 

By capturing the relationship between model uncertainty and knowledge retention, CPCF offers valuable insights into the dynamics of forgetting. Furthermore, the method adheres to the principles of \emph{trustworthy AI} by promoting transparency, reliability, and confidence in decision-making processes.

While the current study focuses on classification, the proposed framework can naturally be extended to regression tasks by transforming CP prediction sets into CP interval estimates. This generalization broadens the scope of the CPCF, making it applicable to a wider range of learning problems while preserving its interpretability and effectiveness.

Future work may focus on optimizing CPCF for greater efficiency and exploring its utility in critical domains such as medical diagnostics, autonomous systems, and cybersecurity, where managing uncertainty and mitigating CF are vital.  
Additionally, extending the approach to handle diverse and complex data distributions, including multimodal, imbalanced, or non-stationary data streams, would further enhance its real-world applicability.

\section*{Acknowledgments}
The work has been supported by the EC through the Horizon Europe/JU SNS project, ROBUST-6G (Grant Agreement no. 101139068). N. Jamoussi is also partially funded by the IMT "Futur, Ruptures \& Impacts" program. 

\bibliographystyle{IEEEtran}
\bibliography{refs}

\end{document}